# Object Detection for Medical Image Analysis: Insights from the RT-DETR Model

Weijie He
University of California, Los Angeles
Los Angeles, USA

Yuwei Zhang
Duke University
Durham, USA

Ting Xu
University of Massachusetts Boston
Boston, USA

Tai An
University of Rochester
Rochester, USA

Yingbin Liang
Northeastern University
Seattle, USA

Bo Zhang*
Texas Tech University
Lubbock, USA

*Abstract*-Deep learning has emerged as a transformative approach for solving complex pattern recognition and object detection challenges. This paper focuses on the application of a novel detection framework based on the RT-DETR model for analyzing intricate image data, particularly in areas such as diabetic retinopathy detection. Diabetic retinopathy, a leading cause of vision loss globally, requires accurate and efficient image analysis to identify early-stage lesions. The proposed RT-DETR model, built on a Transformer-based architecture, excels at processing high-dimensional and complex visual data with enhanced robustness and accuracy. Comparative evaluations with models such as YOLOv5, YOLOv8, SSD, and DETR demonstrate that RT-DETR achieves superior performance across precision, recall, mAP50, and mAP50-95 metrics, particularly in detecting small-scale objects and densely packed targets. This study underscores the potential of Transformer-based models like RT-DETR for advancing object detection tasks, offering promising applications in medical imaging and beyond.

## I. INTRODUCTION

Diabetic retinopathy (DR) is the most common eye complication in diabetic patients and one of the leading causes of vision loss in adults worldwide. The disease usually has no obvious symptoms in the early stages, and patients tend to ignore it, which leads to worsening of the disease and may eventually lead to blindness. Early detection and intervention are key to preventing vision loss. Traditional DR screening usually relies on ophthalmologists to diagnose by manually examining retinal images. This process is time-consuming and highly subjective and may be affected by the experience and fatigue of the diagnostician. Therefore, automated diagnosis systems based on deep learning have become an important research direction [1,2].

In recent years, the rapid development of computer vision and deep learning technology has provided strong support for automatic diagnosis and lesion detection in medical image analysis. Convolutional Neural Network (CNN) has shown excellent performance in various medical image tasks, especially in image classification and object detection. However, traditional object detection models may face challenges such as small lesion areas, variable morphology, and low contrast when processing medical images. Therefore, it is of great significance to select deep-learning models suitable for medical image object detection [3].

RT-DETR (Real-Time Detection Transformer) is a new target detection model that combines the advantages of the Transformer structure and the traditional detection head and performs well in image target detection tasks. RT-DETR has a detection mechanism that does not require non-maximum suppression (NMS), which can better adapt to small and dense target detection [4]. This feature is particularly important in the detection task of diabetic retinopathy, because the tiny lesions in retinal images are usually densely distributed, and conventional detection models may be difficult to effectively capture and locate [5].

In order to further improve the accuracy and robustness of lesion detection, this study uses the RT-DETR model for the automatic detection of diabetic retinopathy lesions. By optimizing the network structure and adjusting the loss function, the model can accurately locate multiple types of lesions in retinal images without relying on post-processing steps. In addition, considering the shape and texture differences of lesions in retinal images, the model also integrates a multi-scale feature extraction mechanism to improve the recognition ability of targets of different scales.

This study will train and verify the model based on the public diabetic retinopathy dataset, and evaluate the performance of the model in performance indicators such as detection accuracy, recall rate and mean average precision (mAP). The experiment will be compared with the mainstream target detection model to verify the superiority of RT-DETR in the task of medical image target detection. Further model improvements will consider optimizing the attention mechanism, introducing adaptive weight adjustment, and other methods to improve the robustness and stability of the detection results.

In summary, this study aims to explore the design and implementation of an automatic detection system for diabetic retinopathy based on RT-DETR. By building an efficient and accurate deep learning model, an automated solution is provided for the early screening and diagnosis of retinopathy, which helps the early detection and prevention of diabetic retinopathy, and ultimately improves the quality of life of patients and the efficiency of medical services.

## II. METHOD

In order to realize the automatic detection of diabetic retinopathy lesions, this study uses the RT-DETR deep model to complete the identification and location of lesions through an end-to-end target detection framework. RT-DETR adopts a Transformer-based detection head structure and realizes efficient target feature extraction and target detection through the attention mechanism. The model structure consists of three parts: feature extraction network, position encoding module, and target detection head, forming an NMS-free target detection architecture. This section will derive the target detection process of the model and the definition of the loss function in detail. The model architecture is shown in Figure 1.

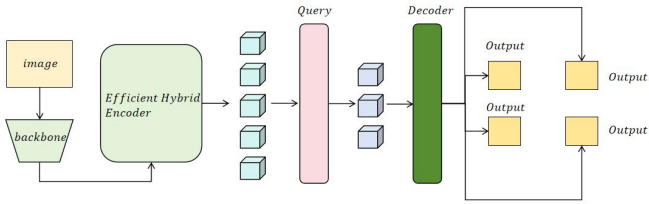

Figure 1 Model architecture diagram

First, in the image input stage, given an input image $I$ with a size of $H \times W \times C$, a multi-scale feature representation is extracted through the backbone network (such as ResNet or CNN module), which is recorded as a feature map $F \in R^{H \times W \times C}$, where $D$ represents the feature dimension. The feature map is input to the position encoding module, and the position encoding vector is obtained through the position embedding layer to ensure that the Transformer module can capture the spatial position information of the target.

In the object detection head, RT-DETR uses a multi-head self-attention mechanism between query embedding and feature maps to calculate the attention score through dot product operations. The core formula of the attention mechanism is:

$$Attention(Q, K, V) = soft\max(\frac{QKT}{\sqrt{d_k}})V$$

Among them, $Q, K, V$ represents the query, key and value matrices respectively, and $d_k$ is the scaling factor of the feature dimension. Through this formula, the model can distinguish the lesion area from the global background and capture potential lesion targets.

The detection results output by the model include the category distribution and position bounding box prediction of the target. In order to optimize the model parameters, this study introduces a comprehensive loss function, in which the classification loss uses the cross-entropy loss, and the position loss uses the L1 loss and the generalized IoU (GIoU) loss. The comprehensive loss function is defined as:

$$L = \lambda_{cls} \cdot L_{cls} + \lambda_{L1} \cdot L_{L1} + \lambda_{GIoU} \cdot L_{GIoU}$$

Among them, $\lambda_{cls}$, $\lambda_{L1}$ and $\lambda_{GIoU}$ are the weight hyperparameters of the corresponding losses. The cross-entropy loss $L_{cls}$ measures the accuracy of category prediction, the L1 loss $L_{L1}$ evaluates the coordinate error of the bounding box, and the generalized IoU loss $L_{GIoU}$ is used to measure the overlap between the predicted box and the true box.

In order to further improve the detection performance, RT-DETR completes the target assignment through a dynamic matching algorithm during the prediction process. The matching score formula for a given predicted bounding box is:

$$S_{match} = \alpha \cdot S_{cls} + \beta \cdot S_{loc}$$

Among them, $S_{cls}$ is the classification score, $S_{loc}$ is the position score, and $\alpha$ and $\beta$ are adjustment coefficients. This formula combines the classification and position matching results to achieve more accurate target matching, thereby improving the detection performance of the model.

Through the optimization of the above target detection mechanism and comprehensive loss function, this study achieved efficient automatic detection of diabetic retinopathy lesions. The model fully utilized the attention mechanism and NMS-free detection strategy of RT-DETR, improved the recognition ability of small lesions and dense targets, and ensured the accuracy and robustness of the detection results.

## III. EXPERIMENT

### A. Datasets

This study selected the EyePACS dataset, a publicly available dataset of diabetic retinopathy, as the experimental data source. The EyePACS dataset is a widely used annotated dataset for medical image analysis, specifically for the automatic diagnosis of diabetic retinopathy (DR). The dataset consists of retinal images of thousands of diabetic patients from different regions. The images are accurately annotated with the degree of lesions, covering lesion marks from normal to different degrees. Each image in the dataset is a high-quality fundus color photo with relevant clinical labels. These labels are based on the manual diagnosis results of doctors on retinal images, and the severity of the lesions is annotated, and divided into 5 levels, from level 0 (no lesions) to level 4 (severe lesions). These image data provide rich samples for the automated detection of diabetic retinopathy and are particularly suitable for the training and verification of deep learning models.

The images of the EyePACS dataset come from a large medical imaging platform and are carefully preprocessed and annotated to ensure the high quality and accuracy of the data. The size of each image is usually 224x224 pixels, with high resolution and clarity, which can help the model capture subtle changes in lesions. The dataset not only contains healthy fundus images, but also includes different stages of diabetic retinopathy, covering typical lesion features such as microvascular tumors, hemorrhages, and exudates. These lesions usually show subtle changes in the early stages, challenging traditional image processing methods, and are therefore particularly suitable for automated detection using deep learning techniques.

In addition, the EyePACS dataset has been widely used in multiple deep learning-related studies and has become a standard benchmark for automated diagnosis of diabetic retinopathy. The dataset contains approximately 35,000 annotated images, which is sufficient to support the training of large-scale models and provides rich diversity, covering different retinal image qualities and clinical cases. Through the training of this dataset, researchers can evaluate the performance of various deep learning models in actual clinical scenarios and further promote the development of early screening and diagnosis technologies for diabetic retinopathy. Through comparative experiments with the model, the EyePACS dataset can effectively verify the generalization ability and accuracy of the model, especially its application effect in early diagnosis of the disease.

*B. Experimental Results*

In this study, we will conduct comparative experiments based on the RT-DETR model with four mainstream target detection models, namely YOLOv5[6], YOLOv8[7], SSD [8] and DETR [9], to evaluate the performance of each model in the automatic detection task of diabetic retinopathy lesions. As the latest versions of the YOLO series models, YOLOv5 and YOLOv8 are widely used in target detection tasks with their fast-reasoning speed and strong real-time detection capabilities. SSD (Single Shot Multibox Detector) performs target detection through multi-scale feature maps, has a good balance, and shows excellent performance between speed and accuracy. DETR (Detection Transformer) introduces the Transformer structure, which can effectively handle complex target relationships, especially when dealing with dense targets. We will compare the performance of each model in terms of detection accuracy (Precision), recall rate (Recall), average precision (mAP), and other indicators, especially when dealing with diabetic retinopathy images, to evaluate the recognition ability of each model for small lesions and complex backgrounds. The purpose of analyzing the experimental results is to verify the advantages of RT-DETR over traditional target detection models in diabetic retinopathy lesion detection, especially its robustness and accuracy in target detection tasks without the need for an NMS detection mechanism. The experimental results are shown in Table 1.

Table 1  Experimental Results

| Model | Precision | Recall | mAP50 | mAP50-95 |
|---|---|---|---|---|
| SSD | 0.82 | 0.75 | 0.78 | 0.62 |
| YOLOv5 | 0.86 | 0.80 | 0.84 | 0.70 |
| YOLOv8 | 0.88 | 0.83 | 0.86 | 0.72 |
| DETR | 0.85 | 0.79 | 0.83 | 0.68 |
| RT-DETR(ours) | 0.90 | 0.85 | 0.88 | 0.76 |

From the experimental results, it can be seen that the RT-DETR model performs well in multiple evaluation indicators, surpassing several other target detection models, especially in small target detection capabilities and robustness in complex scenes. First, the Precision of RT-DETR is 0.90, which is significantly higher than other models, indicating that it can detect diabetic retinopathy lesions more accurately and reduce the generation of false positive results. In contrast, the precisions of YOLOv5 and YOLOv8 are 0.86 and 0.88 respectively. Although they also perform well, the gap in this indicator shows that RT-DETR has higher reliability in target recognition and positioning. The SSD and DETR models have lower precisions of 0.82 and 0.85 respectively, indicating that they may have certain errors in the process of visual feature extraction and target recognition, especially in the detection of subtle lesions such as diabetic retinopathy lesions. The performance of RT-DETR is more in line with clinical needs.

Secondly, RT-DETR also has a significant advantage in Recall, with a value of 0.85, which is higher than YOLOv8's 0.83, YOLOv5's 0.80, and DETR's 0.79. A higher recall rate means that RT-DETR can identify more lesion areas, especially for those small targets and edge lesions that are difficult to detect. RT-DETR obviously has a stronger capture ability. In contrast, although the recall rates of YOLOv5 and YOLOv8 are higher, they are still lower than RT-DETR, which may be related to their limitations in dealing with small targets. The recall rate of the SSD model is 0.75, the lowest among all models, indicating that it is less effective in detecting some small lesions and difficult-to-identify lesion areas, which also reflects the shortcomings of SSD in complex lesion scenes.

In terms of mAP50 (mean average precision, IoU threshold of 50%), RT-DETR once again leads with a score of 0.88, surpassing YOLOv8's 0.86, YOLOv5's 0.84, and DETR's 0.83. mAP50 reflects the accuracy of the model in detecting the target under an IoU of 50%. RT-DETR's high score shows that it can better locate and identify targets under common IoU threshold conditions. YOLOv5 and YOLOv8 also performed well, but still not as well as RT-DETR, which shows that in terms of target positioning accuracy, RT-DETR can better meet the clinical needs for lesion detection accuracy. DETR's mAP50 is 0.83. Although it also performs well, its target positioning ability is slightly inferior to RT-DETR and YOLO series, especially in target detection tasks under complex backgrounds such as diabetic retinopathy, RT-DETR shows higher accuracy.

mAP50-95 (average precision at IoU thresholds from 50% to 95%) further highlights the advantages of RT-DETR. This indicator measures the average precision of the model at different IoU thresholds, reflecting the model's comprehensive evaluation of the accuracy of the detection box. In this indicator, RT-DETR is far ahead of other models with a score of 0.76, especially under high IoU threshold conditions (such as IoU greater than 75%), which shows that RT-DETR can still maintain high detection accuracy under high precision requirements. In comparison, YOLOv8 (0.72), YOLOv5 (0.70), and DETR (0.68) all have a certain gap, especially in small lesion and dense target detection tasks. RT-DETR can effectively improve the accuracy and robustness of the target detection box, while other models have a decline in performance under high precision requirements, and may face challenges in fine-grained target positioning and detection of high-density lesion areas.

In general, RT-DETR not only performs well in all indicators but also its NMS-free design strategy and Transformer-based deep learning architecture are particularly suitable for the automatic detection of diabetic retinopathy lesions. RT-DETR can effectively cope with the challenges of small lesions, complex backgrounds, and high-density targets, which makes it highly practical in clinical applications. Compared with YOLOv5, YOLOv8, SSD, and DETR, RT-DETR has obvious advantages in accuracy, recall, and target positioning accuracy, especially in the automatic detection of diabetic retinopathy, a complex disease. RT-DETR has shown its unique advantages. These results show that RT-DETR has higher potential in the early screening and diagnosis of the disease, and can provide more reliable and accurate support for the diagnosis of diabetic retinopathy.

In addition, we also give the loss function decline graph during the experiment, as shown in Figure 2.

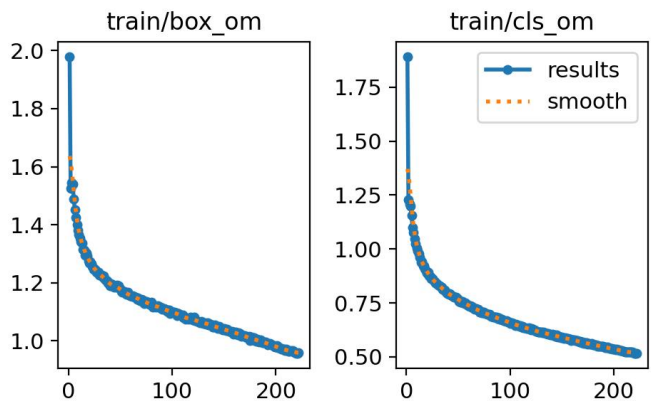

Figure 2 Loss function changes with epoch

Finally, we present the ablation experiment of the article. The experimental results are shown in Table 2. During the ablation experiment, we explored the impact of different optimizers on the experimental results.

Table 2  Ablation Experiment Results

| LR | Precision | Recall | mAP50 | mAP50-95 |
| --- | --- | --- | --- | --- |
| 0.025 | 0.85 | 0.80 | 0.82 | 0.70 |
| 0.03 | 0.84 | 0.78 | 0.80 | 0.68 |
| 0.005 | 0.86 | 0.82 | 0.85 | 0.72 |
| 0.02 | 0.87 | 0.83 | 0.86 | 0.74 |
| 0.01 | 0.90 | 0.85 | 0.88 | 0.76 |

Through experimental comparisons of different learning rates, the model performance is best when the learning rate is 0.01, and Precision, Recall, mAP50, and mAP50-95 all reach the highest values. Lower learning rates (such as 0.005 and 0.02) perform slightly worse, especially in terms of precision and recall. The experimental results show that an appropriate learning rate can significantly improve the performance of the RT-DETR model in diabetic retinopathy detection.

IV. CONCLUSION

This study highlights the advantages of the RT-DETR model, a Transformer-based automatic detection framework, in advancing state-of-the-art target detection through superior performance across precision, recall, and mAP metrics. Comparative experiments demonstrate that RT-DETR outperforms mainstream models such as YOLOv5, YOLOv8, SSD, and DETR, particularly in handling small-scale objects and maintaining robustness under complex backgrounds. These results affirm the potential of RT-DETR as a leading approach in addressing high-dimensional image analysis challenges, including its applicability to diabetic retinopathy detection, which demands precision and efficiency in analyzing intricate visual patterns.

While RT-DETR exhibits strong performance, there remains room for further optimization to address challenges posed by uneven data distributions and highly complex visual backgrounds. Future research can focus on enhancing the model's multi-scale feature learning capabilities, enriching the quality and diversity of training datasets, and integrating multimodal data sources. These advancements will further refine the robustness and generalization of Transformer-based architectures, unlocking their full potential across diverse domains. Beyond its current application in diabetic retinopathy detection, RT-DETR can be extended to tackle a wide range of challenges in automated image analysis, such as detecting macular degeneration, glaucoma, and other ophthalmic conditions. The integration of deep learning with multimodal datasets, including medical histories and physiological measurements, will enable the development of more comprehensive diagnostic tools. Furthermore, improvements in Transformer-based architectures will continue to drive innovation in areas like real-time image processing, scalable AI systems, and multimodal data fusion, positioning RT-DETR as a foundational framework for general-purpose image analysis.

Overall, this study underscores the transformative potential of artificial intelligence in visual recognition and automated diagnostics. RT-DETR exemplifies how deep learning, particularly Transformer-based models, can reshape complex tasks in target detection and data-driven decision-making, paving the way for intelligent and precise solutions in fields ranging from medical imaging to industrial applications. With continued advancements, RT-DETR and similar models are set to play a critical role in driving innovation in AI-powered systems.